\documentclass[runningheads]{llncs}
\usepackage{graphicx}
\usepackage[cmex10]{amsmath}
\usepackage{amssymb}
\usepackage{amsfonts}
\usepackage{algorithm}
\usepackage{algpseudocode}
\usepackage{amsthm}
\usepackage{graphicx}
\usepackage[caption=false,font=footnotesize]{subfig}
\usepackage{epsfig}
\usepackage{cite}
\usepackage{tensor}
\usepackage{color}
\usepackage{gensymb}
\usepackage{multirow}
\usepackage{comment}
\usepackage{blindtext}
\usepackage{multicol}
\usepackage{graphicx}
\usepackage{caption}

\graphicspath{{figures/}}
\DeclareGraphicsExtensions{.eps}

\theoremstyle{definition}

\DeclareMathOperator*{\argmin}{arg\,min}

\begin{document}
\title{A Self-Organizing Network with Varying Density Structure for Characterizing Sensorimotor Transformations in Robotic Systems%
\thanks{This work is supported in part by the Research Grants Council (RGC) of Hong Kong under grant number 14203917, and in part by PROCORE-France/Hong Kong Joint Research Scheme sponsored by the RGC and the Consulate General of France in Hong Kong under grant F-PolyU503/18.}
}
\titlerunning{A Varying Density Self-Organizing Network}
\author{Omar Zahra \and David Navarro-Alarcon}
\authorrunning{Zahra and Navarro-Alarcon}%
\institute{Department of Mechanical Engineering, The Hong Kong Polytechnic University, Hung Hom, Kowloon, Hong Kong.
\email{dnavar@polyu.edu.hk}\\
}
\maketitle              
\begin{abstract}
In this work, we present the development of a neuro-inspired approach for characterizing sensorimotor relations in robotic systems.
The proposed method has self-organizing and associative properties that enable it to autonomously obtain these relations without any prior knowledge of either the motor (e.g. mechanical structure) or perceptual (e.g. sensor calibration) models. Self-organizing topographic properties are used to build both sensory and motor maps, then the associative properties rule the stability and accuracy of the emerging connections between these maps. 
Compared to previous works, our method introduces a new varying density self-organizing map (VDSOM) that controls the concentration of nodes in regions with large transformation errors without affecting much the computational time.
A distortion metric is measured to achieve a self-tuning sensorimotor model that adapts to changes in either motor or sensory models.
The obtained sensorimotor maps prove to have less error than conventional self-organizing methods and potential for further development.

\end{abstract}
\section{Introduction}
Among the many interesting cognitive abilities of animals and humans is the motor babbling process that leads to the formation of the sensorimotor map. 
Many theories have been introduced about how these behaviors develop since prenatal stages \cite{zoia2007evidence}. 
This sensorimotor adaptation paradigm has proven to be useful in robotics for relating motor commands with sensory outputs when prior/exact knowledge of its model is unavailable (which is common in practice). 
A robot with changes to its mechanical structure (e.g. due to damage) or perceptual system (e.g. due to sensor decalibration) is generally not able to properly coordinate its motions without updating such sensorimotor relations. 
Drawing inspiration from the adaptive properties of living organisms, artificial neural systems can be developed to cope with these uncertainties. 
The development of computational sensorimotor models with adaptation properties can lead to the emergence of valuable self-calibrating behaviors. 
Additionally, these could help to (safely) verify theories about the internal workings of the human brain, but with machines.

Previous studies with primates have concluded a topographic arrangement in areas dedicated to motor and sensory processing, where adjacent body parts tend to have an adjacent representation in the brain cortex \cite{kaas1997topographic,silver2009topographic}. 

Thus, to represent perceptual computing units in a biologically inspired manner, such topologically preserving property was considered.

Topographic models are useful for characterizing sensory and motor spaces in robots. 
Yet, to co-relate how a particular motion/configuration produces a sensory stimuli, additional associative properties must be considered.
One common model for linking different brain areas based on shared activity patterns is the so-called Hebbian rule \cite{hebb2002organization}. 
It states that if two neuronal regions are persistently activated together, the connection between them is strengthened; the connection is weakened if no simultaneous activity is present. 
Topographic and associative properties are the basis for the sensorimotor adaptive method that we propose in this paper.

In the literature, many efforts have been placed to model human sensorimotor abilities with methods based on self-organizing maps (SOM) \cite{schillaci2014online,kumar2010ctrlvisual,buessler1999multiple}. 
Most of these works use SOMs as a topography-preserving and dimension-reducing tool, to map several sensor readings with motor actions.

In \cite{kumar2010ctrlvisual}, an SOM is used to form a sensory map with visual feed. However, the learning process to form a sensorimotor map takes place mainly through gradient-descent rule which makes it less biologically plausible.
In \cite{schillaci2014online}, two dynamic SOMs (DSOM) \cite{rougier2011dynamic} representing head and arm of a humanoid robot were used to achieve visuo-motor coordination. Yet, that model suffered from a degradation in performance when perturbations were added to motor commands.
In \cite{buessler1999multiple}, the sensorimotor coordination is achieved by utilizing bi-directional neural modularity such that motor output can be predicted from sensory input and vice versa. 
For the proposed method in this paper, the learning paradigm allows for the development of reciprocal correlations inherently while maintaining high accuracy.

In this study, we propose a new method for representing sensorimotor transformations of robotic systems.
The neuro-inspired method combines self-organizing and associative properties to model continuously adapting relations between sensory and motor spaces.
Compared to previous works, our new method proposes a varying density SOM (VDSOM) that reduces the transformation error that is typically present at the periphery of standard SOMs.
This is done by automatically adjusting a parameter that controls the density of neighboring nodes at regions with large transformation errors . In case of changes in either motor or sensory models, a distortion metric is measured to readjust the formed sensorimotor map to suit these changes.
The resulting computational model can effectively reduce the mean error over the whole map, while coping with changes in the original sensorimotor model.
Several cases of study (such as transformation accuracy, amputation, limb extension) are presented to thoroughly evaluate the proposed method.

The rest of this paper is organized as follows: Section \ref{sec:methods} describes the computational model; Section \ref{sec:results} presents its quantitative evaluation; Section \ref{sec:conclusions} gives final conclusions.

\section{Methods}\label{sec:methods}
\subsection{A Biologically-Inspired Sensorimotor Model}

Human bodies have different morphologies which develop over years (from birth to death) and even subject to drastic changes as in the case of amputations. 
However, the brain somehow always manages to find or re-adapt such mappings between sensory feedback and motor actions. 
In infants, for example, motor babbling helps to adaptively obtain these sensorimotor relations, where by performing motions covering the workspace, the brain is able to correlate bodily configurations with its corresponding motor actions \cite{piaget1952origins}\cite{von1982eye}. 

It is also clear from recent studies that in both sensory and motor areas in the brain, adjacent body parts have also contiguous representations \cite{penfield1937somatic}. Moreover, many of these areas are connected together by some synapses which develop connections based on their joint activity. Among these rules is the well-known Hebbian learning rule \cite{hebb2002organization}. 

To represent such learning paradigm, a model for human sensorimotor mapping is constructed using SOM (modeling topographically arranged brain areas) and Hebbian learning rules (modeling connections among these areas).
Both of these models have clear biologically-inspired properties as they can represent topographic organization of neurons and modulation of strength of synaptic connections, respectively.

SOM are built upon the underlying rules of development of cognitive functions, as they encode competition, cooperation and adaptation \cite{Book:Kohonen2001}.
The nodes (neurons) of the SOM compete against each other such that only one becomes the best matching neuron (BMU) for a given input.
However not only the BMU contribute to the output, but also the neighboring neurons as well, such that the closer to the BMU the greater would be the contribution to the output. 
This represents the lateral interaction between neurons in a network. 
Adaptation by modulation of weights of BMU (and neighborhood nodes) occurs to enhance the chance of the BMU to represent the input vector and act as the BMU again for a similar input.

The Hebbian learning rule wires the SOMs representing the sensory space and motor space together, such that neurons active on both sides at the same time have the strength of the synaptic connection in between increased proportional to the magnitude of activity of both the pre-synaptic and post-synaptic neurons. 
These connections achieve sensorimotor correlation between the motor actions and the corresponding sensory input that happen to be active at the same time. 

\begin{figure}
	\centering
	\includegraphics[scale=.35]{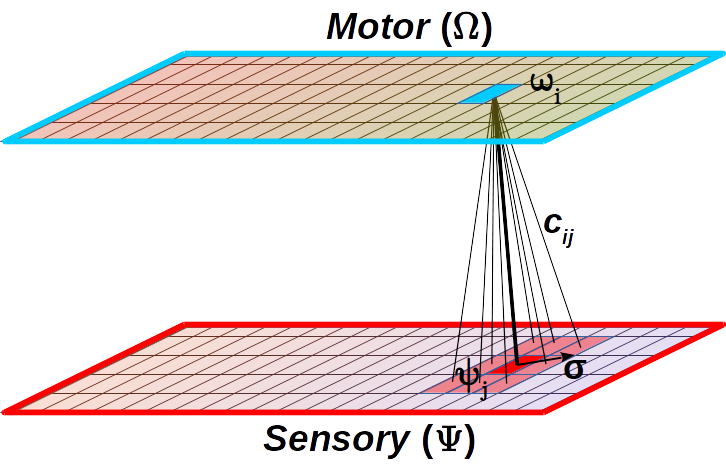}
	\caption{Motor space ($\Omega$ connected to Sensory space ($\Psi$) through Hebbian connections (\textbf{\textit{$c_{ij}$}}). As learning process proceeds, active nodes in $\Omega$ ($\omega_{i}$) have the connections to active ones in $\Psi$ reinforced ($\psi_{j}$ and neighborhood within a radius of $\sigma$).}
	\label{fig:cs}
\end{figure}

\vspace{-4.0mm}
\subsection{Modeling Sensory and Motor Spaces}
The SOM is formed of a 2 dimensional lattice of $M$ neurons (nodes), each of them associated with a weight vector ($w_{i}$) of dimension as each vector in the input space ($X$). 
These weights are initially set to random values, then, data training points are introduced in a random fashion to the SOM. 
When a vector of data $x$ is introduced to the SOM, the node with least Euclidean distance between its weights and the input vector is chosen to be the best matching unit (winning neuron) based on:
\begin{equation}
i = \argmin_j \Vert w_j - x \Vert
\end{equation}
where $i$ the denotes the index of the BMU.
The weights of all the neurons in the neighborhood around the BMU are updated to give a closer approximation of the input vector $x$.
This node is computed with the following update rule:
\begin{equation}
w_{j}(t+1) = w_{j}(t) + \alpha(t) h_{ji}(t) (x - w_{j}(t))
\end{equation}
where $h_{ji}$ is the neighboring function, which is computed with the following Gaussian function:
\begin{equation}
h_{ji}(t) = \exp\left(\dfrac{-\Vert{r_{j}-r_{i}  \Vert^{2}}}{2\sigma^{2}(t)}\right)
\end{equation}
where $r_{i}$ and $r_{j}$ are the positions of the BMU and the neighboring $j$th node within the lattice, respectively. 
The learning rate $\alpha$ and neighborhood radius $\sigma$ are set to decrease exponentially with time such that:
\begin{align}
\left .\sigma(t) = \sigma_{init} \exp\left(\dfrac{-t}{T}\right) ,    
\right. \alpha(t) = \alpha_{init} \exp\left(\dfrac{-t}{T}\right)
\end{align}
where $t$ is the time of current iteration, $T$ is the desired time constant of decrease, $\alpha_{init}$ and $\sigma_{init}$ are the initial values of the learning rate and neighborhood radius, respectively. 
By tunning the adequate parameters for the learning process, the weights of the nodes are updated to give an adequate mapping for both sensory and motor states within the identified robot workspace.

\subsection{Formation of Sensorimotor Mapping}
To provide a correlation between activity of each node in motor space $a_{i}$ to sensory space $a_{j}$ back and forth as shown in Fig.\ref{fig:cs}, the Hebbian Oja learning rule \cite{oja1982simplified} is applied by applying the equation:
\begin{equation}
\textit{c}_{ij}(t+1) =  \textit{c}_{ij}(t) + \eta \left(a_{i} a_{j}-\textit{c}_{ij}(t) a_{j}^{2} \right)
\label{eqn:hebb_rule}
\end{equation}
where ${c}_{ij}$ represents the strength of the connection between the pre and post synaptic nodes, while $\eta$ is the learning rate. Nodes from both maps that happen to be active at the same time tend to have high correlations and thus stronger synaptic connection between these nodes. The first term in the parenthesis ensures applying Hebbian learning rule to achieve the correlation. 
The second term guarantees the stability of the learning process where a forgetting term is included such that in case some nodes are not active for a long time the strength of the connection is attenuated. 

The activity $a_j$ of each node is calculated by applying the following Gaussian kernel for the Euclidean distance between the weights of the nodes and the input vector:
\begin{equation}
a_j(t) = \exp \left(\dfrac{-\Vert w_j(t)-x \Vert^{2}}{\sigma^{2}(t)} \right)
\end{equation}
This expression gives rise of a one-to-one mapping between the nodes of the two SOMs (that respectively model the motor and sensory spaces). The resulting connections are reciprocal (i.e. bidirectional). This means that they can be used to either predict the sensory states based on a given motor action, or to compute the required motor actions to achieve a certain sensory state \cite{Saegusa2009}.

\subsection{Varying Density Structure}
The sensorimotor mappings can be achieved by combining SOM and Oja-Hebbian learning rules, as described above. 
However, the naive use these method results in regions (e.g. the periphery of the lattice) with large transformation errors.
Two initial hypothesis were assumed to cause this problem. 
The first is that having a small number of training points at these regions may cause that problem. 
The second one is that having comparatively low number of neurons near the boundaries to represent the sensorimotor correlations may the culprit (e.g. having fewer neurons affect the accuracy of the estimated values). Such problem at the boundaries is one of the drawbacks of the SOM mentioned in the literature \cite{kohonen2013essentials}.

For the former hypothesis, training data with higher density at the lattice boundaries was used, however, it did not improve the mapping accuracy. 
A viable solution was to increase the density of the neurons near the problematic regions such as the boundaries of both the sensory and motor maps to give a better representation at these points. 
To achieve this behavior, the SOM update rule was modified by proposing a different neighborhood function that gives the required variable density of nodes. 
This is done by calculating the summation of the norm of the weights of the BMU to each node in the lattice then applying the Gaussian function.
The \emph{node density coefficient} $\rho$ is computed as follows:
\begin{equation}
\rho = \exp \left(- \sum_{i\in O} \Vert w_{bmu}-w_{i} \Vert^{2} \right)
\end{equation}
for $O$ as the local neighborhood surrounding the neuron.
This function aims to give a smooth gradient effect of contribution of proximal nodes. 

The coefficient $\rho$ can be used to quantitatively determine neurons with a small number of neighbors.
More neurons can be attracted to these nodes to have a denser population and therefore give a better approximation of corresponding values in the sensorimotor map. 
The resulting map is characterized by having a variable density (even when using uniform training data) that controls the number of nodes in a region based on $\rho$; we call this network a varying density SOM (VDSOM).
The additional term $\rho$ shall have a minimal effect in the formation of the network at the beginning and increase as the learning process proceeds. 
On the other hand, if it increases at a slow rate the exponential decay term of the neighborhood radius would make the effect of that term minimal. 

To achieve this effect, the new neighborhood is defined as follows:
\begin{equation}
h(t) =  {\left(\dfrac{t}{\rho T}\right)}^{4} \exp\left(\dfrac{-t}{\sigma^{2}(t)T}\right)
\end{equation}
where the new term was chosen to be of the fourth order to have adequate values without disturbing the dynamics of the learning process.

By adding that term, the lattices formed for both the sensory and motor spaces are more dense at the boundaries. 
This helps to reduce the transformation errors that occurs in these regions without the need to increase the total number of neurons in the network. 
This density regulation concept may not (yet) have some proof from a neuro-biological perspective, however, varying densities of neurons is certainly present in many different areas of the brain and within each area. 
For example, in the primary visual cortex, the central region has a higher density of neurons relative to the the peripheral regions. 
In most primates, the central vision area is the main region of interest when observing a scene \cite{collins2010neuron}.
Additionally, the proposed mechanism to automatically increase the number of neurons agrees with studies in which high neuronal density is observed for processing hand and face fine motions \cite{young2013cell}.
Although this study focuses on VDSOM, the same concept can be applied to vary the structure of a Growing Neural Gas(GNG) network \cite{fritzke1995growing}  to obtain the optimal number of nodes to represent the same sensorimotor model.

\subsection{Adaptation to Changes in the Sensorimotor Model}

Note that in case of changes in body morphology (e.g.generated by attaching of an external limb or amputation) or changes in the perceptual system (e.g. by wearing vision inverting goggles \cite{Journals:Kohler1962}) the computed sensorimotor model is no longer representative. 
For this situation, both, sensory and motor maps should be updated accordingly, as well as the inter-connections representing the transformations between these spaces. 
However, in traditional SOM, once the learning process reaches the specified number of iterations, changes in input data---corresponding to sensory/motor information---will not modify the networks structure.
This results in a model that no longer adapts, and therefore is not able to represent the new (and actual) sensory/motor configurations.

To overcome this drawback, a distortion metric $\zeta$ is incorporated into the method. 
If the $\zeta$ is found to exceed a give (arbitrary) threshold value after the mapping is established, the neighborhood radius $\sigma$ is reset to an adequate value to be able to re-adapt the network's structure. 
Such distortion metric is computed as:
\begin{equation}
\zeta = \dfrac{1}{n}\sum\limits_{i=1}^n  \sum\limits_{x \epsilon X} \Vert x-w_{i} \Vert ^{2}
\label{eqn:dm}
\end{equation}
where $n$ is the number of data vectors $x$ available in the data set $X$. 
The new neighborhood radius $\sigma_{r}$ is set to be initially equal to $\sigma(\tau)$, when the distortion metric after the perturbations occur is equal to that. 
Then, the value of $\sigma_{r}(t)$ can be calculated from the equation:
\begin{equation}
\sigma_{r}(t) = \sigma_{init} exp\left(\dfrac{-(t+\tau)}{T}\right)
\end{equation}

If the value of distortion after perturbations is higher than that at the beginning of the learning process, then the radius is set to the maximum value which is the radius of the SOM.
On the other hand, a modified version of Oja-Hebbian connections is used to adapt better to these changes.
\begin{equation}
\textit{c}_{ij}(t+1) =  \textit{c}_{ij}(t) + \eta(a_{i}a_{j}-\beta\textit{c}_{ij}(t)a_{j}^{2})
\label{eqn:oja_hebb_rule}
\end{equation}
The additional term allows to control the $\textit{forgetting rate}$ of the already formed connections. 

\begin{table}[b!]
	\caption{Mean and Maximum errors for forward and inverse mappings using SOM and VDSOM.}\label{table:error}
	\centering
	\begin{tabular}{|l|l|l|l|l|}
		\hline
		\multirow{2}{*}{Error} & X (mm) & Y (mm) & $\theta_{1}$ ($\degree$) & $\theta_{2}$ ($\degree$) \\ \cline{2-5} 
		& Mean(Max)        & Mean(Max)        & Mean(Max)       & Mean(Max)       \\ \hline
		SOM                    & 2.6(26.0)        & 2.9(29.0)        & 0.37(3.45)      & 0.61(6.70)      \\ \hline
		VDSOM           & 1.15(11.7)       & 1.26(15.0)       & 0.31(2.25)      & 0.44(5.36)      \\ \hline
	\end{tabular}

\end{table}

Thus the values of the additional term $\beta$ and the learning rate $\eta$ are set to allow for new connections to be formed in a faster manner. 
These terms are assigned high values that decrease exponentially based on the following expressions:
\begin{equation}
\beta(t) = \beta_{init} \exp\left(\dfrac{T-t}{T}\right),\quad
\eta(t) = \eta_{init} \exp\left(\dfrac{T-t}{T}\right)
\end{equation}

\section{Results}\label{sec:results}

\subsection{Setup}


A simulation for the computational model of the sensorimotor mapping was built using Tensorflow library\cite{tensorflow2015-whitepaper} on a PC with i7-6500 16GB RAM. Both the system without and with the modifications were simulated for 2D lattice SOMs with square grid of 30x30, 50x50 and 70x70 nodes.

A kinematic model of two link robotic arm was used as the prototype system. 
The end-effector task space is assumed to be measured with an external positions sensor (e.g. a camera).
In our sensorimotor model, the joint space is represented with motor SOM, whereas the task space is represented with a sensory SOM.
Random joint angles within certain ranges were used to generate end-effector positions. 
$L_1$ and $L_2$ denote the lengths of first and second link, respectively, $\theta_{1}$ and $\theta_{2}$ the joint angles of first link relative to the horizontal axis and joint angle of second link relative to the first link. The forward kinematics relation can be simply computed as:
\begin{align}
X = L_1\cos(\theta_{1}) + L_2\cos(\theta_{1}+\theta_{2})\nonumber\\
Y = L_1\sin(\theta_{1}) + L_2\sin(\theta_{1}+\theta_{2})
\end{align}

The connections between both joint space and task space SOMs were developed, as described above, based on the Oja-Hebbian learning rule. 
As can be seen from table \ref{table:error} that both the mean and the maximum error values were drastically reduced after applying the proposed solution to the SOM for forward and inverse mappings. It can also be concluded from table \ref{table:error_2} that as the number of nodes increases the error decreases at the expense of increasing the computational time required to build the network and establish the connections.

\begin{table}[t]
	\caption{Mean and Maximum errors for forward and inverse mappings using VDSOM with different number of nodes.}
	\label{table:error_2}
	\centering
	\begin{tabular}{|l|l|l|l|l|}
		\hline
		\multirow{2}{*}{Error} & X (mm) & Y (mm) & $\theta_{1}$ ($\degree$) & $\theta_{2}$ ($\degree$) \\ \cline{2-5} 
		& Mean(Max)        & Mean(Max)        & Mean(Max)       & Mean(Max)       \\ \hline
		\begin{tabular}[c]{@{}l@{}}VDSOM\\ (30x30)\end{tabular}                    & 4.16(36.3)        & 3.93(21.1)        & 0.91(4.60)      & 1.53(7.62)      \\ \hline
		\begin{tabular}[c]{@{}l@{}}VDSOM\\ (50x50)\end{tabular}                    & 2.13(13.5)        & 2.11(15.3)        & 0.52(2.80)      & 0.80(5.05)      \\ \hline
		\begin{tabular}[c]{@{}l@{}}VDSOM\\ (70x70)\end{tabular}           & 1.15(11.7)       & 1.26(15.0)       & 0.31(2.25)      & 0.44(5.36)      \\ \hline
	\end{tabular}
	
\end{table}

\begin{multicols}{2}
	
	\begin{center}
		\includegraphics[width=0.8\columnwidth,height=0.65\columnwidth]{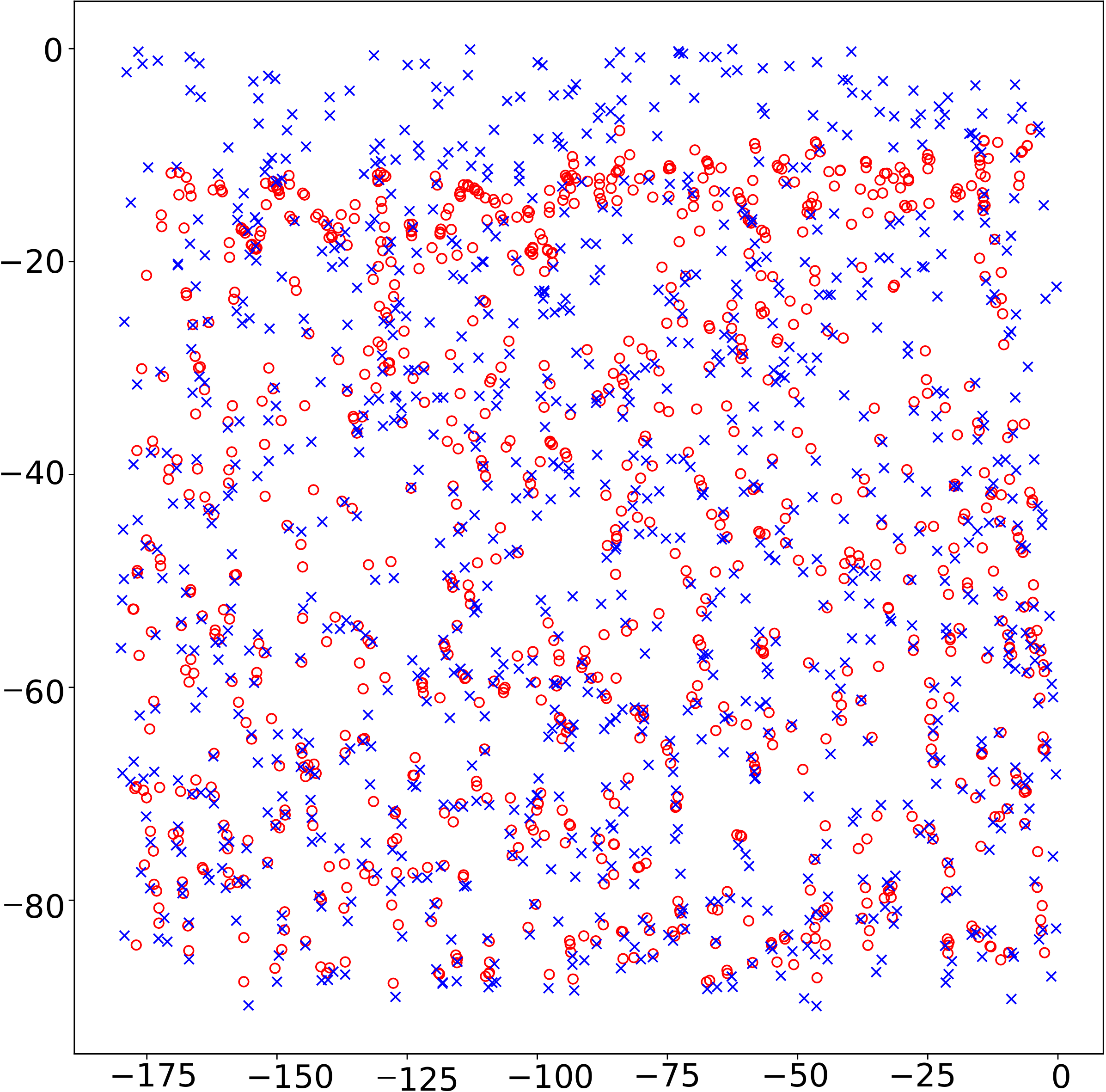}
		\includegraphics[width=0.8\columnwidth,height=0.65\columnwidth]{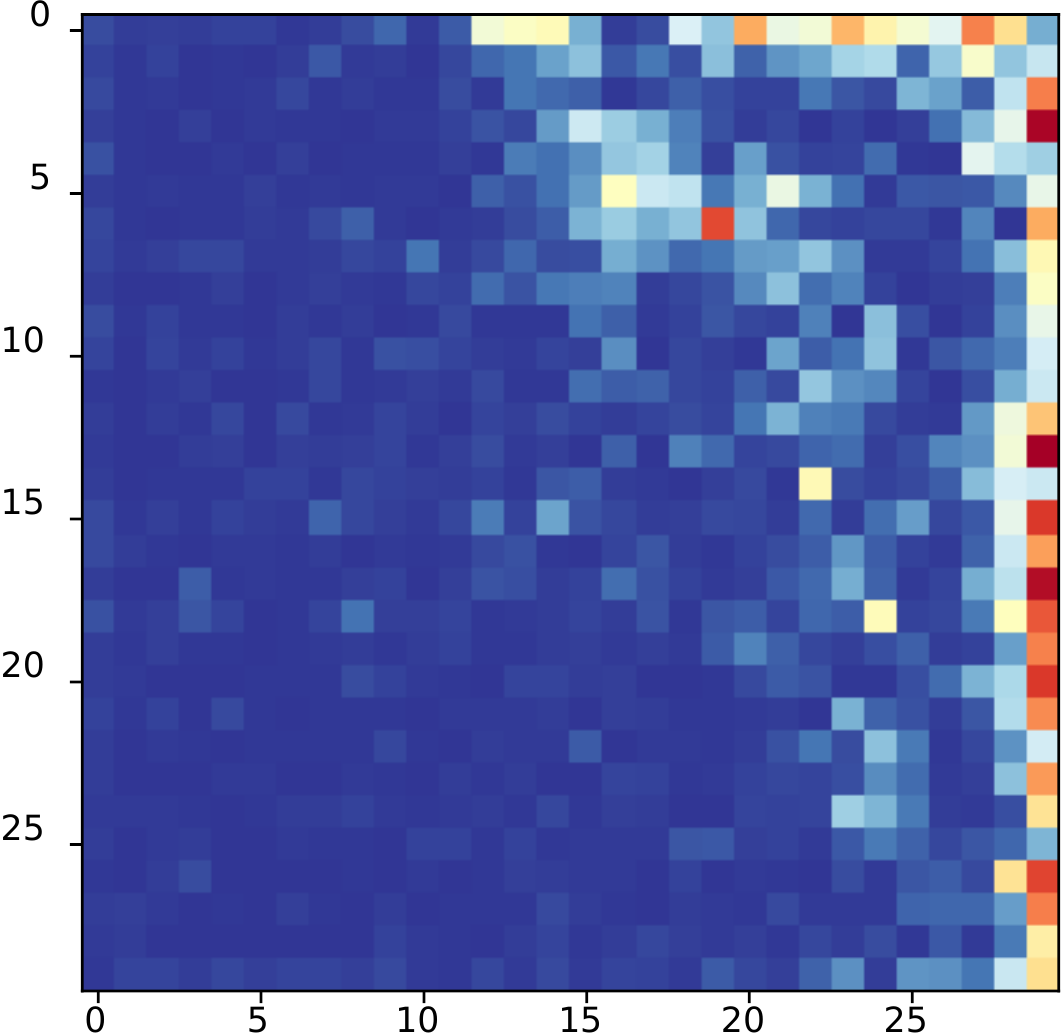}
		\captionof{figure}{Motor SOM with heatmap representing error at each node when it was chosen as a BMU.}
		\label{fig:som_js}
	\end{center}
	
	\begin{center}
		\includegraphics[width=0.8\columnwidth,height=0.65\columnwidth]{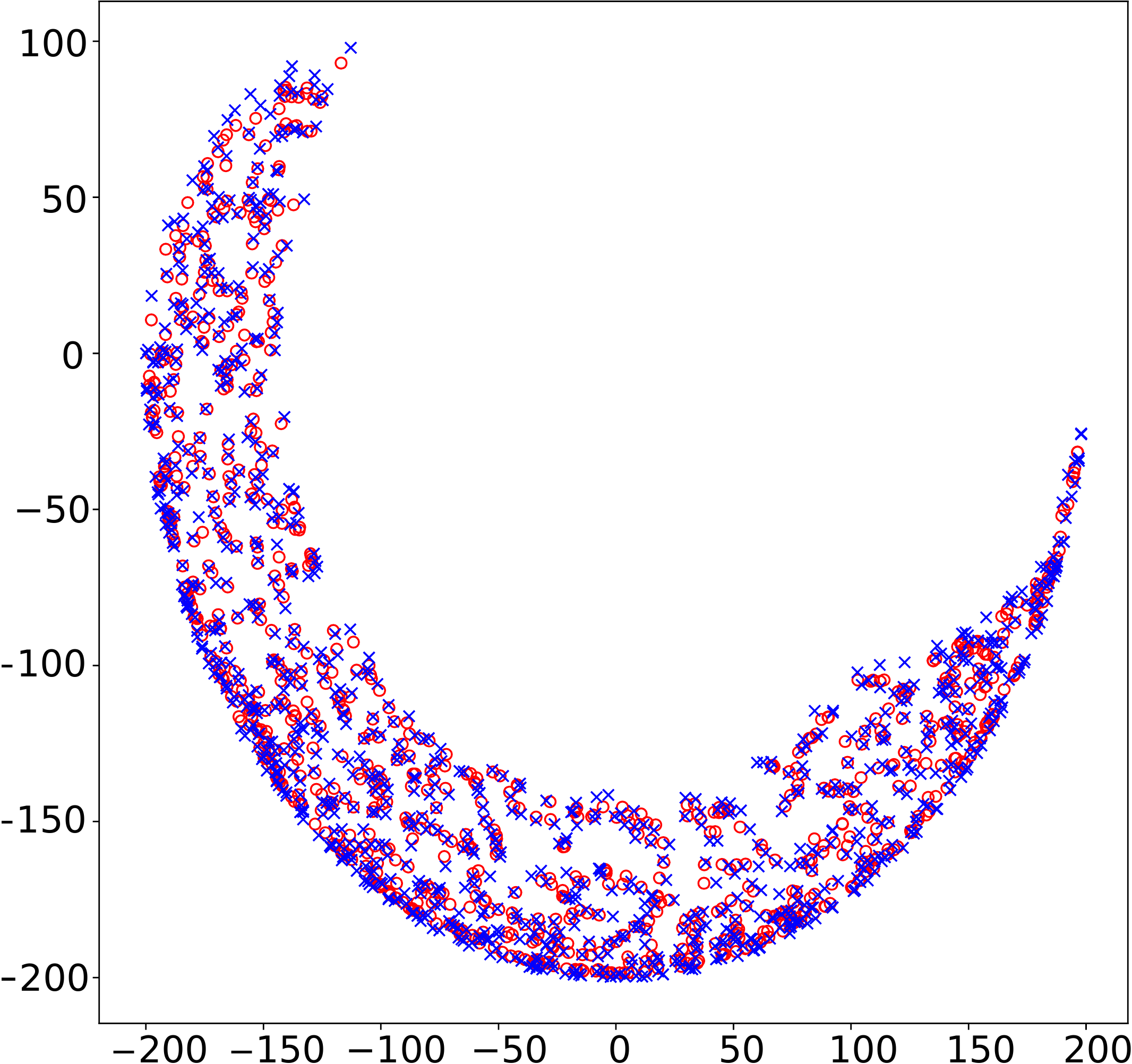}
		\vspace{1.5mm}
		\includegraphics[width=0.77\columnwidth,height=0.63\columnwidth]{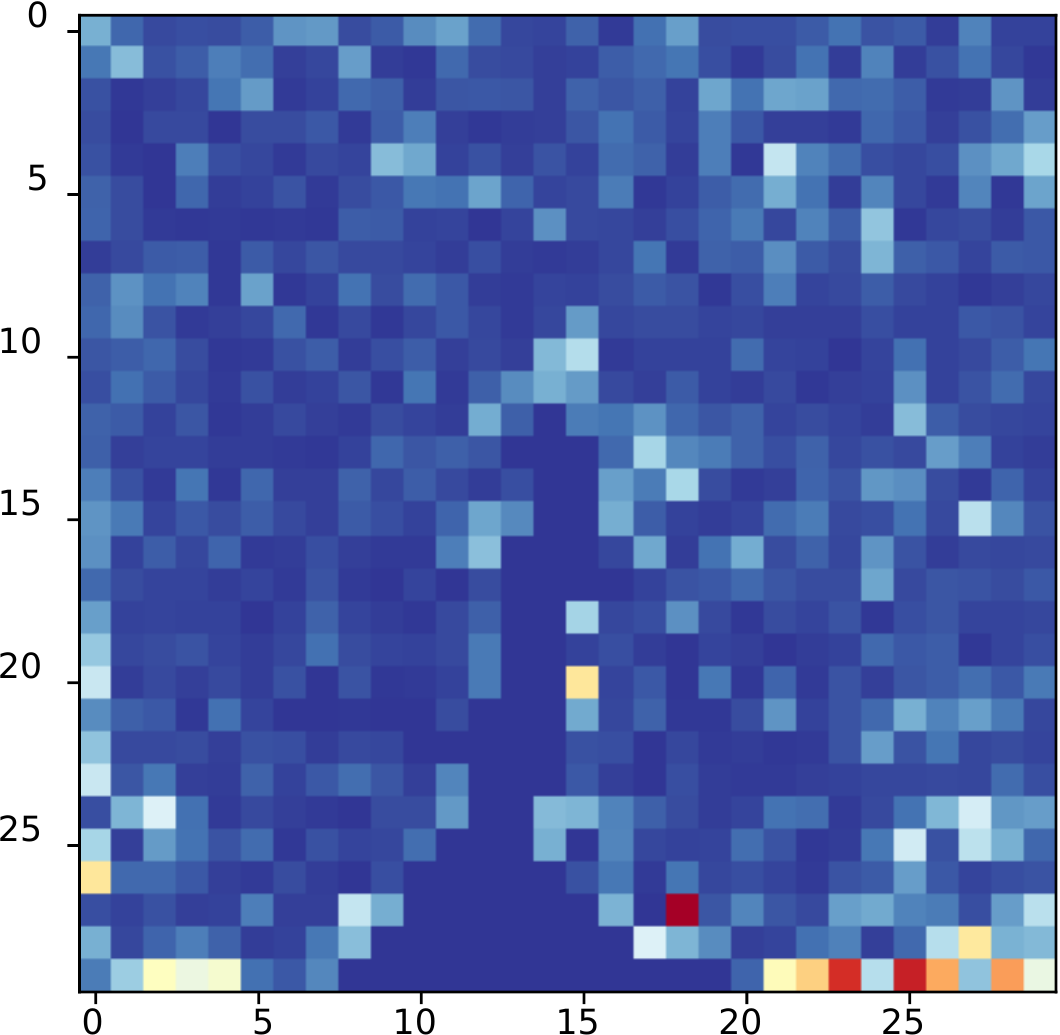}
		\captionof{figure}{Sensory SOM with heatmap representing error at each node when it was chosen as a BMU.}
		\label{fig:som_ts}
	\end{center}
	
\end{multicols}

\begin{multicols}{2}
	
	\begin{center}
		\includegraphics[width=0.8\columnwidth,height=0.65\columnwidth]{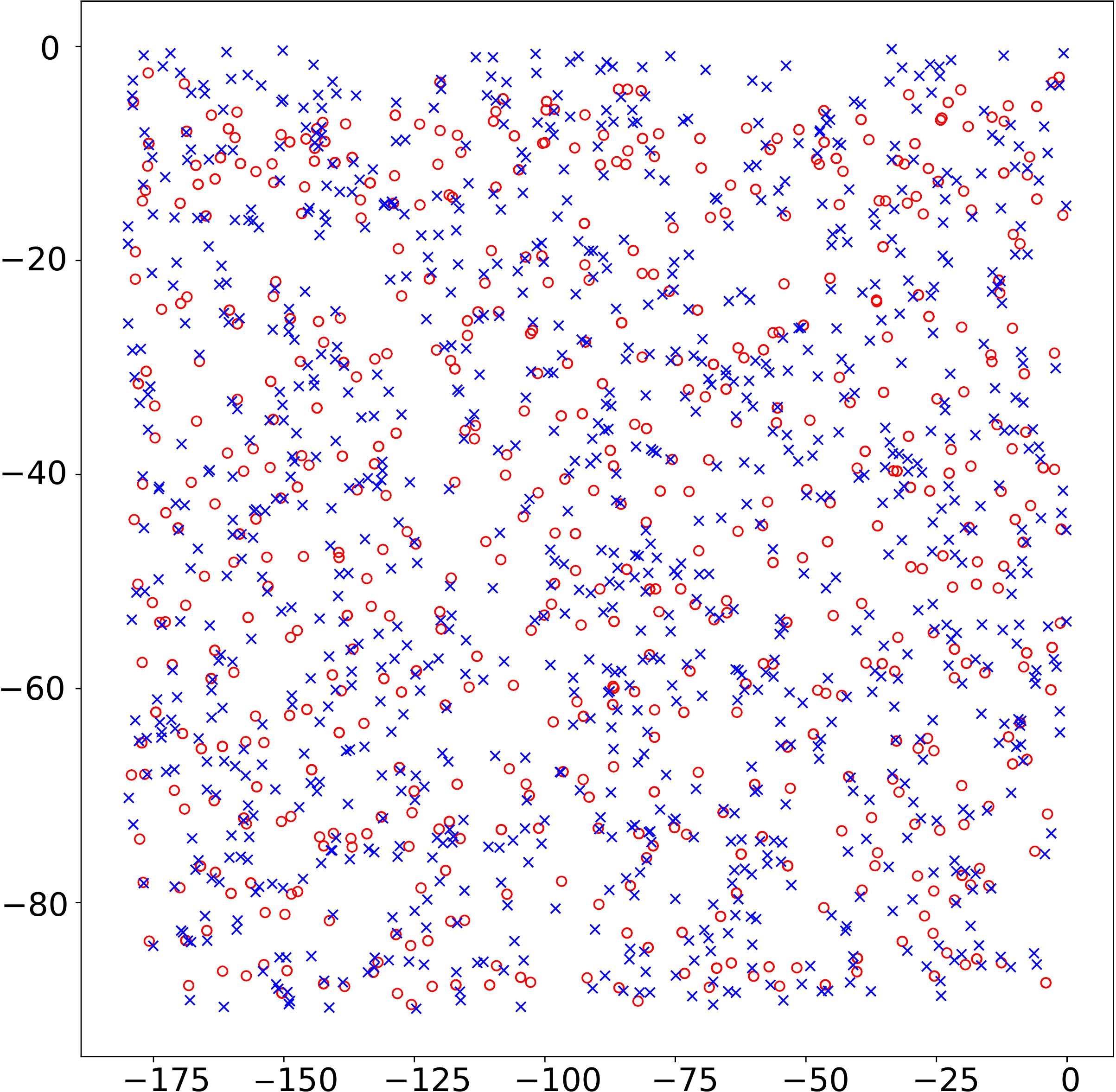}
		\includegraphics[width=0.8\columnwidth,height=0.65\columnwidth]{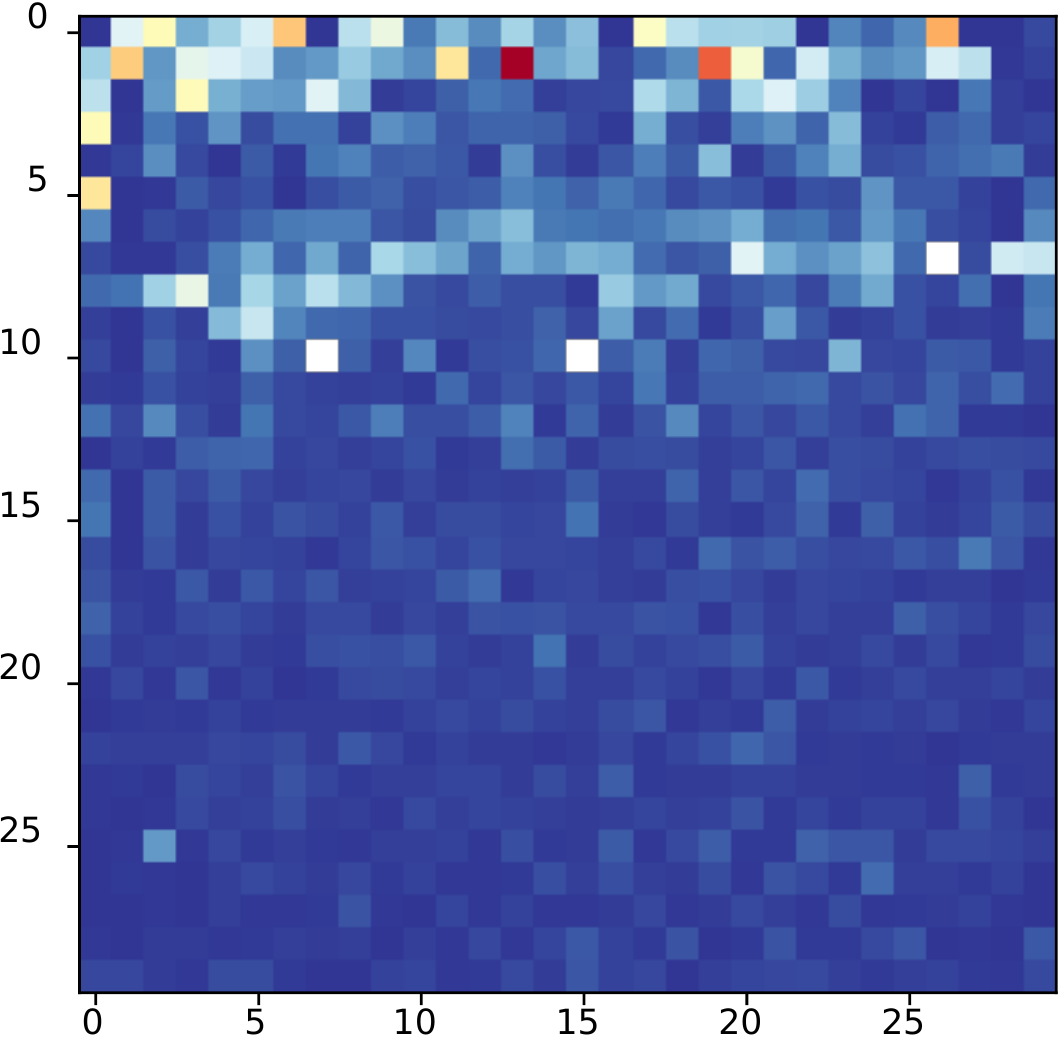}
		\captionof{figure}{Motor VDSOM with error heatmap.}
		\label{fig:vdsom_js}
	\end{center}
	
	\begin{center}
		\includegraphics[width=0.8\columnwidth,height=0.65\columnwidth]{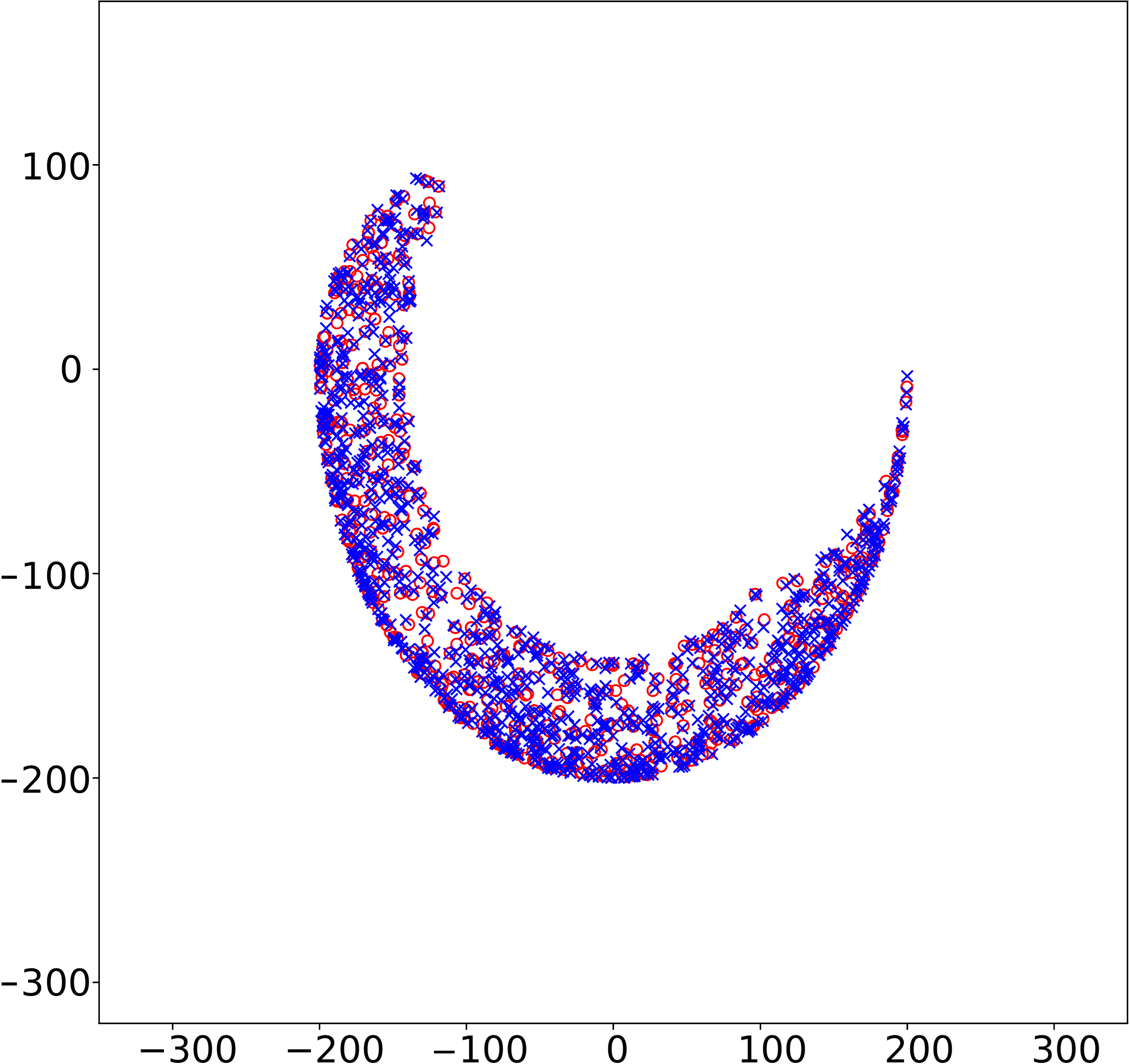}
		\includegraphics[width=0.8\columnwidth,height=0.63\columnwidth]{figures/r_150x150/hm_js_b_.pdf}
		\captionof{figure}{Sensory VDSOM with error heatmap.}
		\label{fig:vdsom_ts}
	\end{center}
\end{multicols}

\subsection{Enhanced Accuracy}
Fig.\ref{fig:som_js} and \ref{fig:som_ts} show the final SOMs developed after running several trials to obtain the most adequate parameters for each SOM. As shown in Fig.\ref{fig:som_ts}, the original SOM covers the whole workspace uniformly but less dense at the peripherals. It can be observed from the error plot in Fig.\ref{fig:som_js} and \ref{fig:som_ts} that higher error values occur at these areas, where the dark blue color and the dark red color represent low error and high error,respectively. The effect of the added factor $\rho$ can be noticed in Fig.\ref{fig:vdsom_js} and \ref{fig:vdsom_ts} where higher density can be observed at the contour of the workspace, and less error in these areas in both forward and inverse mappings. Although the introduced method have an error that is relatively high compared to conventional control methods, it takes one step forward in the formation of biologically-inspired sensorimotor maps.

\subsection{Adaptation to Changes in Morphology}
The robot morphology was altered to simulate attaching and removing a tool from the end effector. To allow the system to detect such changes, the $\zeta$ is calculated based on equation \eqref{eqn:dm} and compared with a threshold
value. Consequently, when such changes are predicted to occur, the learning process is reset to update the mapping. In case of limb length extension, it can be concluded from Fig.\ref{fig:vdsom_ext_sens} and \ref{fig:dm_extend} that the map adapts to re-accommodate for that change and decreases the distortions detected in the computed maps. 
The connections between the sensory and motor maps are updated to represent the new configuration.
Similarly, in the case of limb length reduction, as shown in Fig.\ref{fig:vdsom_amp_mot} and \ref{fig:dm_shortening}, the distortion is measured. 
The change in distortion value, triggers the adaptation mechanism that allows for the maps to be recomputed.

\begin{multicols}{2}
	
	\begin{center}
		\includegraphics[width=0.68\columnwidth,height=0.54\columnwidth]{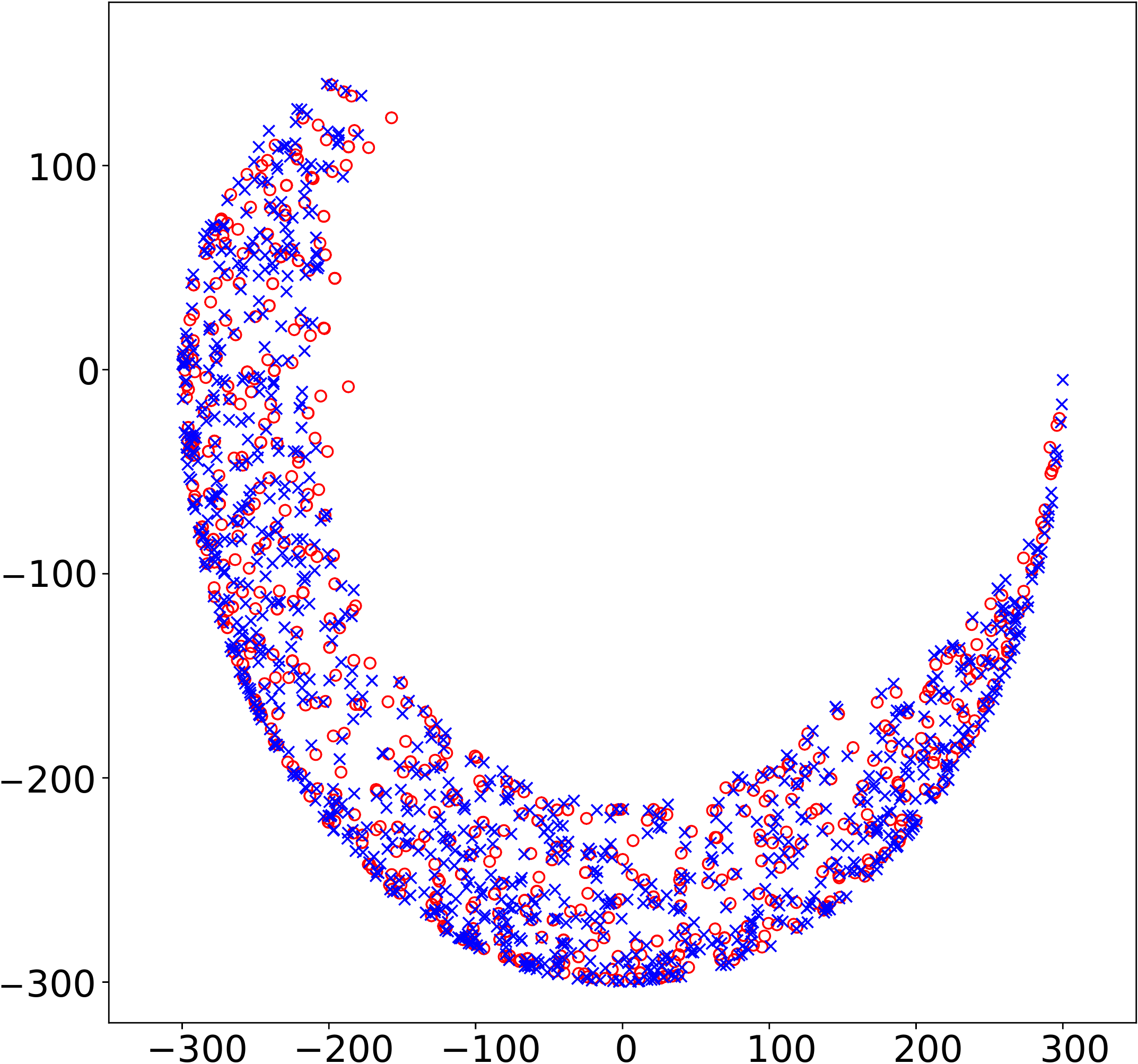}
		\includegraphics[width=0.68\columnwidth,height=0.54\columnwidth]{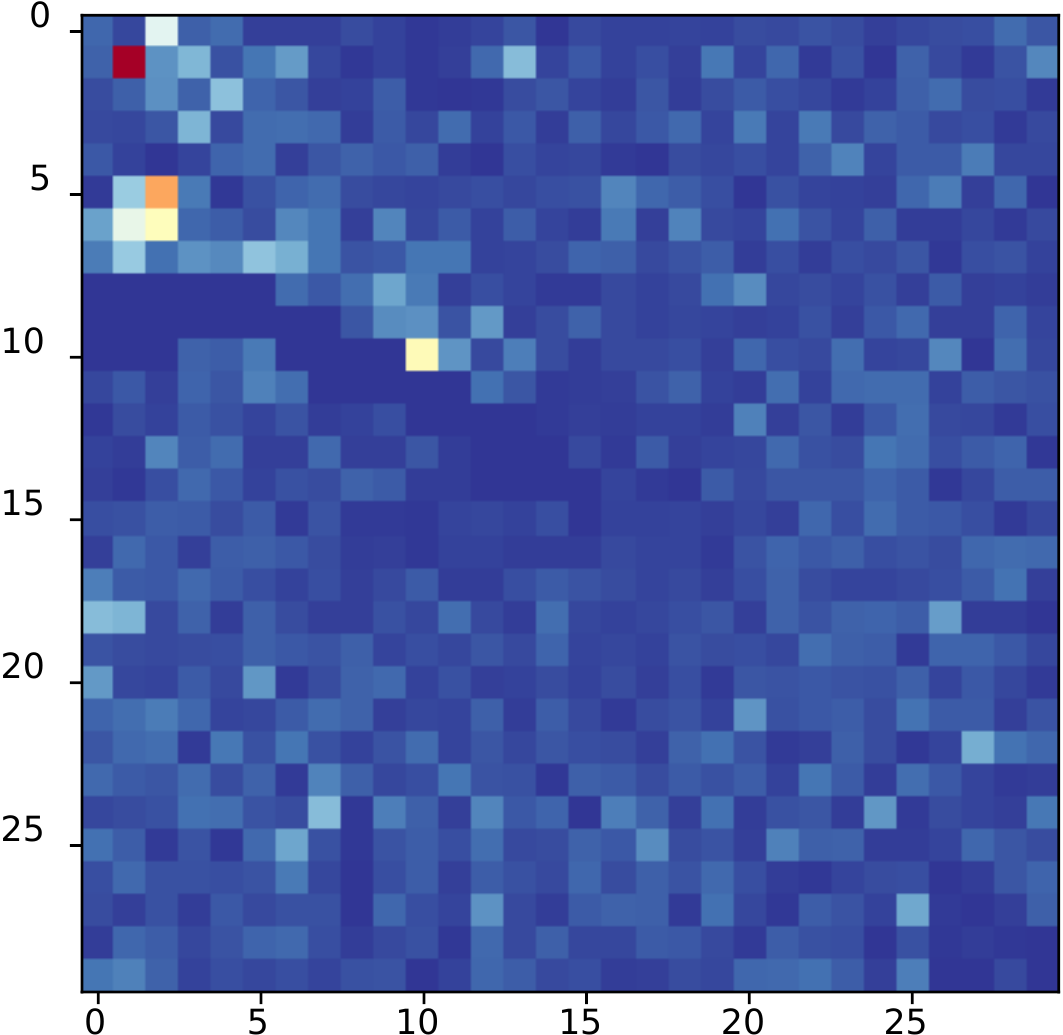}
		\captionof{figure}{Sensory VDSOM after stretching the links with error heatmap.}
		\label{fig:vdsom_ext_sens}
	\end{center}
	
	\begin{center}
		\includegraphics[width=0.68\columnwidth,height=0.54\columnwidth]{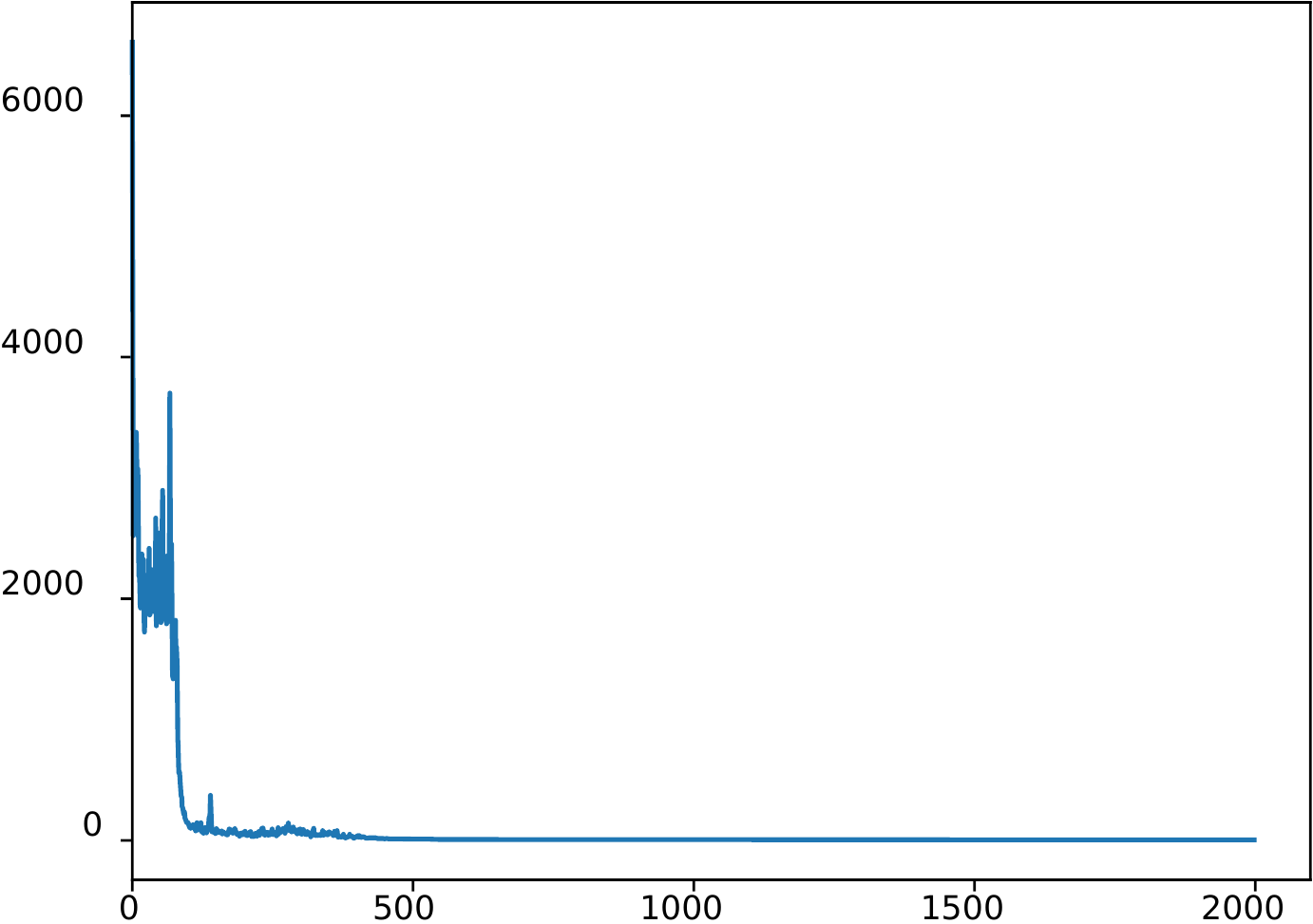}
		\includegraphics[width=0.68\columnwidth,height=0.54\columnwidth]{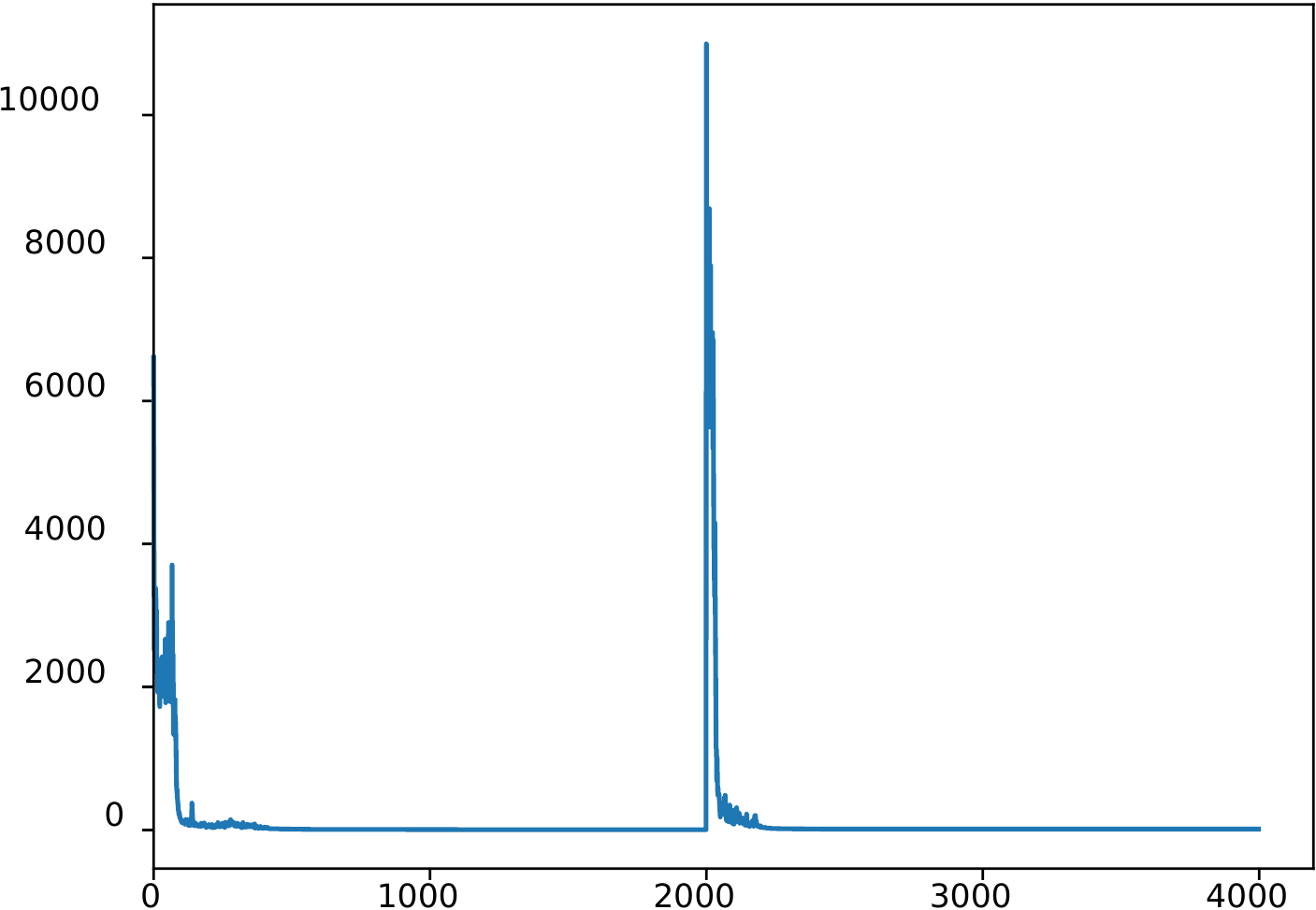}
		\captionof{figure}{Distortion in sensory map before and after stretching the links.}
		\label{fig:dm_extend}
	\end{center}
	
\end{multicols}

\begin{multicols}{2}
	
	\begin{center}
		\includegraphics[width=0.8\columnwidth,height=0.66\columnwidth]{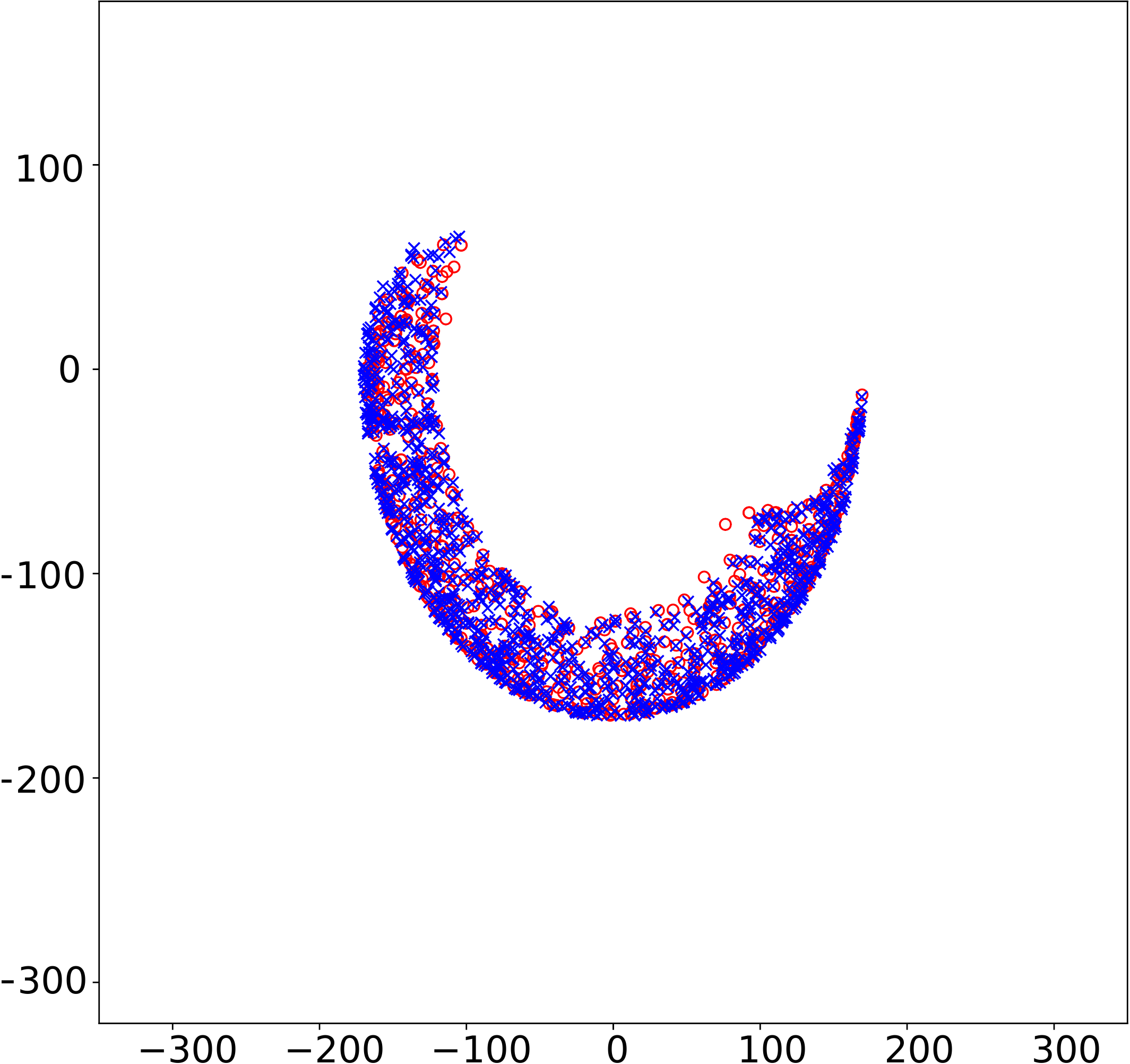}
		\includegraphics[width=0.8\columnwidth,height=0.66\columnwidth]{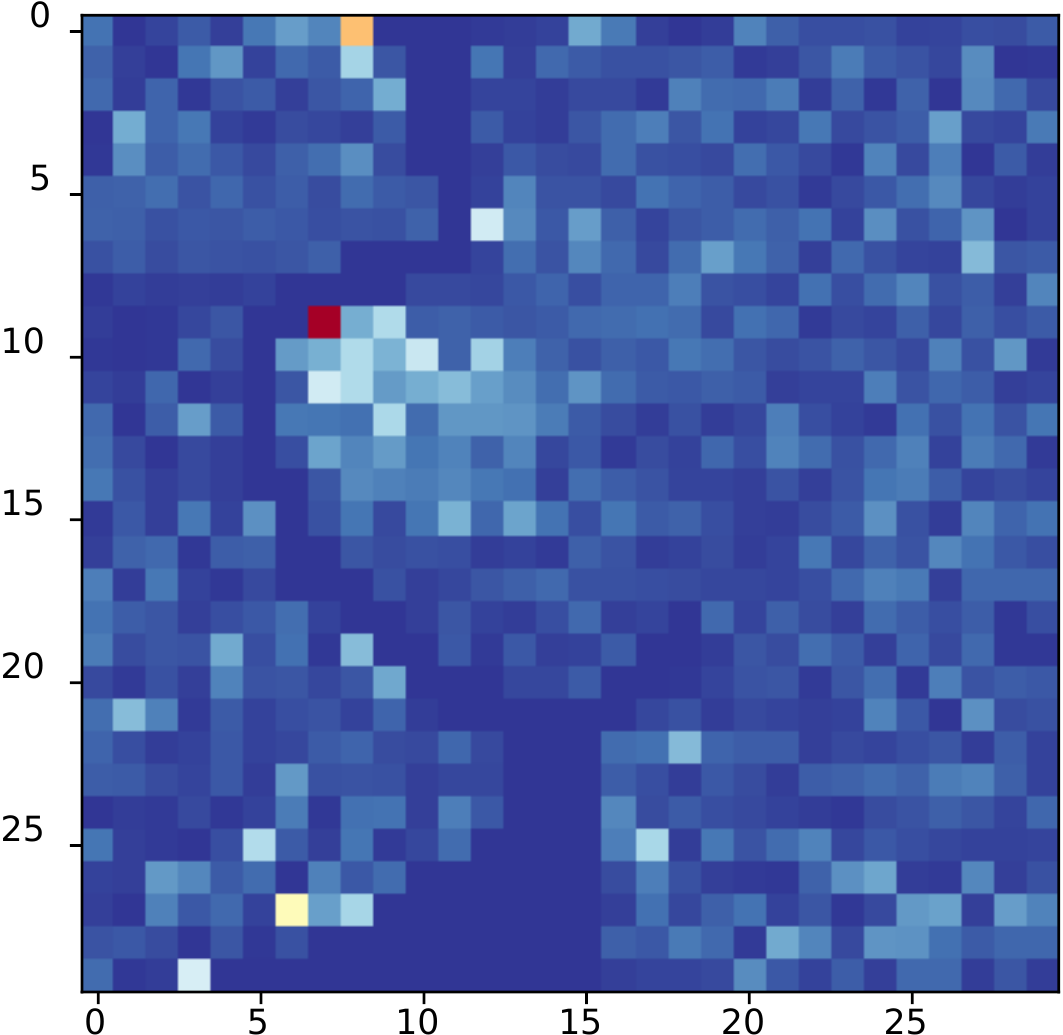}
		\captionof{figure}{Sensory VDSOM after shortening a link with error heatmap.}
		\label{fig:vdsom_amp_mot}
	\end{center}
	
	\begin{center}
		\includegraphics[width=0.8\columnwidth,height=0.66\columnwidth]{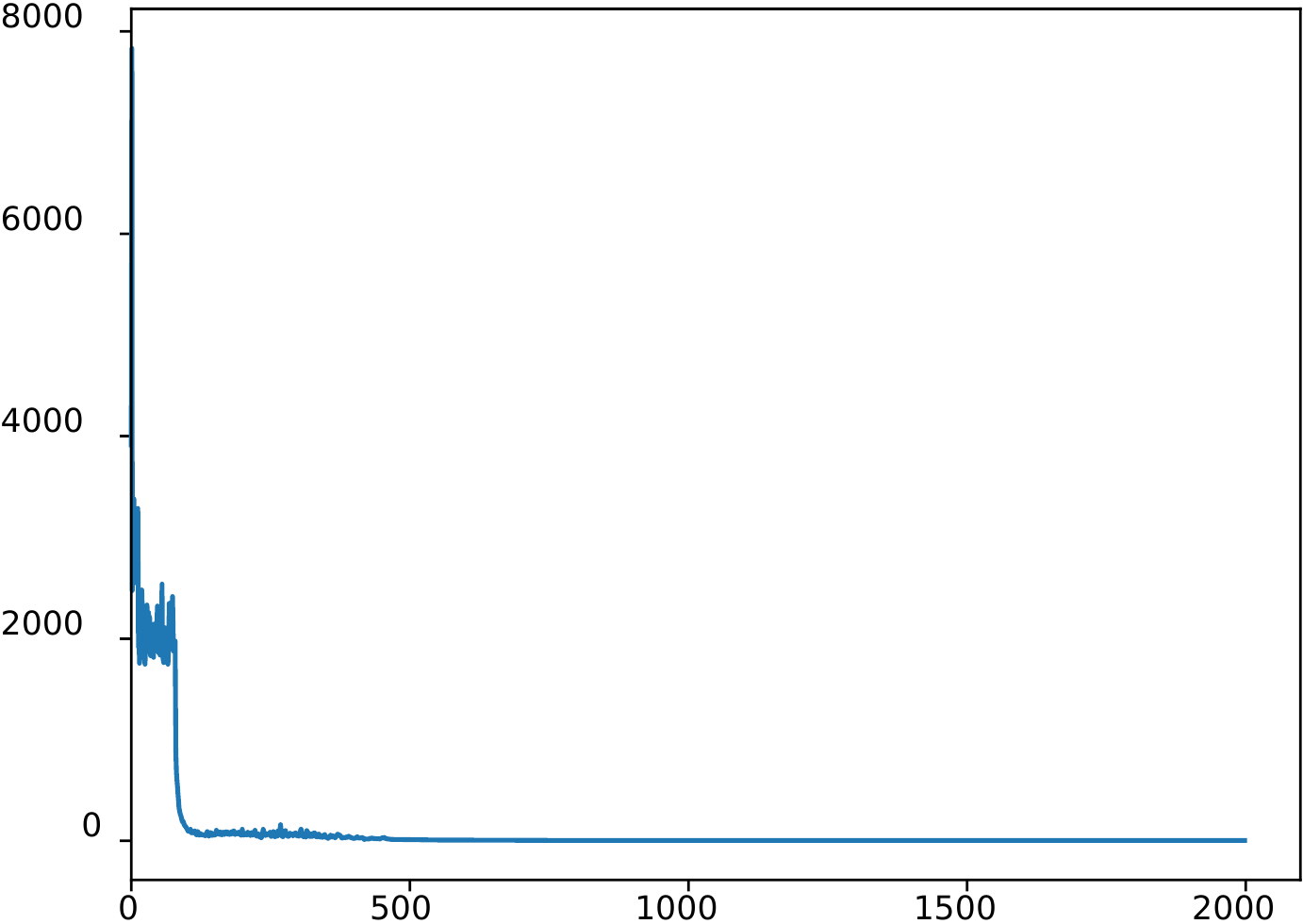}
		\includegraphics[width=0.8\columnwidth,height=0.66\columnwidth]{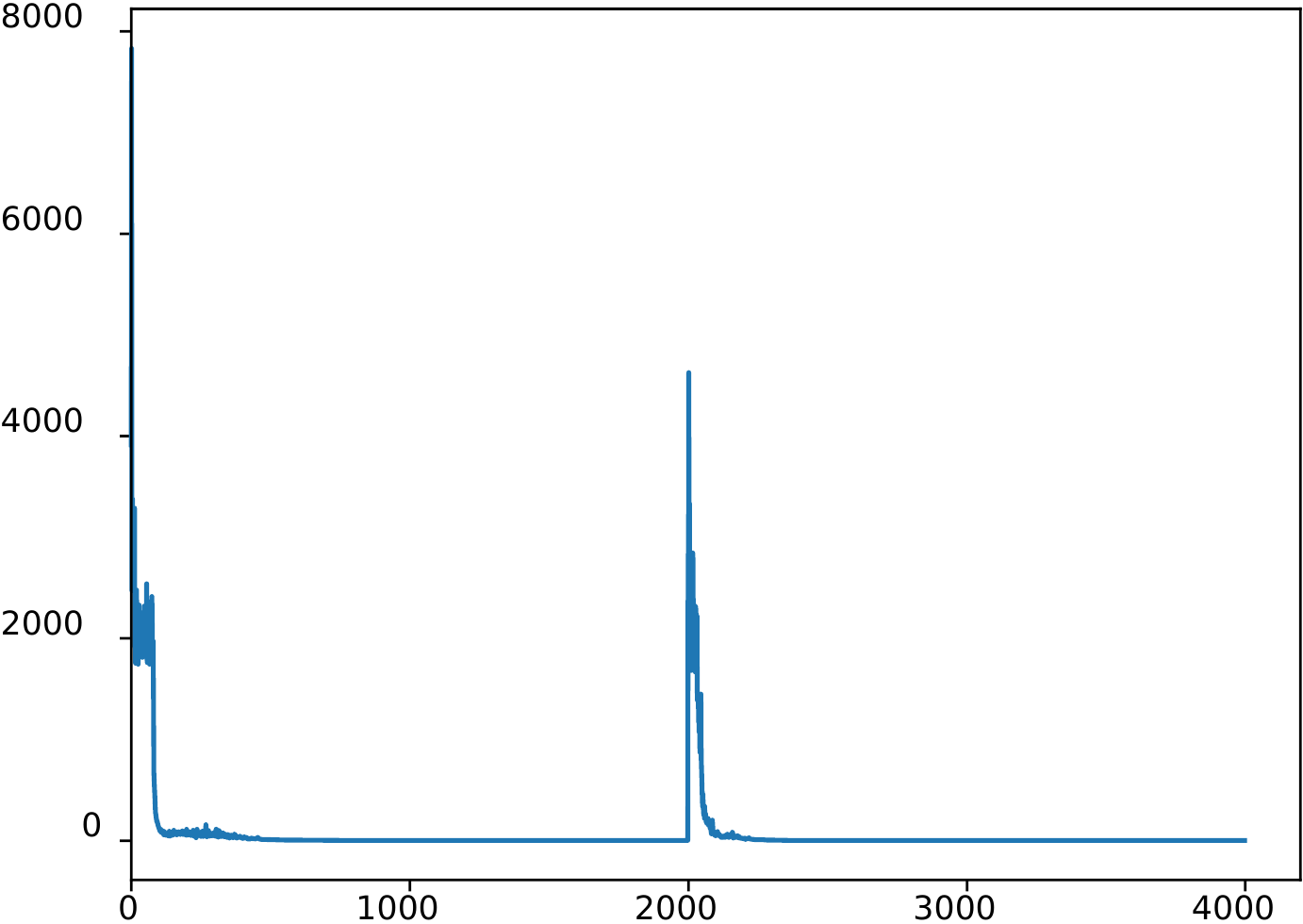}
		\captionof{figure}{Distortion in sensory map before and after shortening the link length.}
		\label{fig:dm_shortening}
	\end{center}
	
\end{multicols}

\section{Conclusions}\label{sec:conclusions}

A sensorimotor map was built to correlate sensory and motor spaces in a discretized form with bidirectional connections. This solution relies on collecting data samples by motor babbling, thus it is adequate to be used for various robotic manipulators without any prior information about robot kinematics. Using the SOM introduced by Kohonen with Oja-Hebbian learning rules the mapping was achieved with noticeable error values at the contour of the SOM -and thus the workspace-. A new neighborhood function was proposed to increase the density of nodes at the contour to give better approximation for the corresponding values. The proposed neighborhood increases the density of the nodes wherever the distance between the weights of the BMU and the neighboring nodes has small values. Finally, a perturbation was introduced to simulate a change in either sensory or motor map. A distortion metric was used to assess the state of the robot and reset the learning parameters to adequate values in case of changes in the morphology. Thus adaption process takes place to update the sensorimotor map, by allowing for changes in both the formed VDSOMs and connections. 

Concerning the current limitations of this method, these maps can only be used for coarse control. A large number of nodes would be needed for fine discretization of the workspace which is computationally inefficient. Additionally, an extended study is needed to utilize the dimension reduction properties of SOM to be fit for robots with higher degrees of freedom.

\bibliography{biblio.bib}

\begin{thebibliography}{10}
\providecommand{\url}[1]{#1}
\csname url@samestyle\endcsname
\providecommand{\newblock}{\relax}
\providecommand{\bibinfo}[2]{#2}
\providecommand{\BIBentrySTDinterwordspacing}{\spaceskip=0pt\relax}
\providecommand{\BIBentryALTinterwordstretchfactor}{4}
\providecommand{\BIBentryALTinterwordspacing}{\spaceskip=\fontdimen2\font plus
\BIBentryALTinterwordstretchfactor\fontdimen3\font minus
  \fontdimen4\font\relax}
\providecommand{\BIBforeignlanguage}[2]{{%
\expandafter\ifx\csname l@#1\endcsname\relax
\typeout{** WARNING: IEEEtran.bst: No hyphenation pattern has been}%
\typeout{** loaded for the language `#1'. Using the pattern for}%
\typeout{** the default language instead.}%
\else
\language=\csname l@#1\endcsname
\fi
#2}}
\providecommand{\BIBdecl}{\relax}
\BIBdecl

\bibitem{zoia2007evidence}
S.~Zoia, L.~Blason, G.~D’Ottavio, M.~Bulgheroni, E.~Pezzetta, A.~Scabar, and
  U.~Castiello, ``Evidence of early development of action planning in the human
  foetus: a kinematic study,'' \emph{Experimental Brain Research}, vol. 176,
  no.~2, pp. 217--226, 2007.

\bibitem{kaas1997topographic}
J.~H. Kaas, ``Topographic maps are fundamental to sensory processing,''
  \emph{Brain research bulletin}, vol.~44, no.~2, pp. 107--112, 1997.

\bibitem{silver2009topographic}
M.~A. Silver and S.~Kastner, ``Topographic maps in human frontal and parietal
  cortex,'' \emph{Trends in cognitive sciences}, vol.~13, no.~11, pp. 488--495,
  2009.

\bibitem{hebb2002organization}
D.~Hebb, ``The organization of behavior. 1949,'' \emph{New York Wiely}, vol.~2,
  no.~7, p.~8, 2002.

\bibitem{schillaci2014online}
G.~Schillaci, V.~V. Hafner, and B.~Lara, ``Online learning of visuo-motor
  coordination in a humanoid robot. a biologically inspired model,'' in
  \emph{Development and Learning and Epigenetic Robotics (ICDL-Epirob), 2014
  Joint IEEE International Conferences on}.\hskip 1em plus 0.5em minus
  0.4em\relax IEEE, 2014, pp. 130--136.

\bibitem{kumar2010ctrlvisual}
S.~Kumar, P.~Premkumar, A.~Dutta, and L.~Behera, ``Visual motor control of a
  7dof redundant manipulator using redundancy preserving learning network,''
  \emph{Robotica}, vol.~28, no.~6, pp. 795--810, 2010.

\bibitem{buessler1999multiple}
J.~Buessler, R.~Kara, P.~Wira, H.~Kihl, and J.~Urban, ``Multiple
  self-organizing maps to facilitate the learning of visuo-motor
  correlations,'' in \emph{IEEE International Conference on Systems Man and
  Cybernetics}, vol.~3, 1999, pp. III--470.

\bibitem{rougier2011dynamic}
N.~Rougier and Y.~Boniface, ``Dynamic self-organising map,''
  \emph{Neurocomputing}, vol.~74, no.~11, pp. 1840--1847, 2011.

\bibitem{piaget1952origins}
J.~Piaget and M.~T. Cook, ``The origins of intelligence in children.'' 1952.

\bibitem{von1982eye}
C.~Von~Hofsten, ``Eye--hand coordination in the newborn.'' \emph{Developmental
  psychology}, vol.~18, no.~3, p. 450, 1982.

\bibitem{penfield1937somatic}
W.~Penfield and E.~Boldrey, ``Somatic motor and sensory representation in the
  cerebral cortex of man as studied by electrical stimulation,'' \emph{Brain},
  vol.~60, no.~4, pp. 389--443, 1937.

\bibitem{Book:Kohonen2001}
T.~Kohonen, \emph{Self-Organizing Maps}, ser. Information Sciences.\hskip 1em
  plus 0.5em minus 0.4em\relax Springer Berlin Heidelberg, 2001.

\bibitem{oja1982simplified}
E.~Oja, ``Simplified neuron model as a principal component analyzer,''
  \emph{Journal of mathematical biology}, vol.~15, no.~3, pp. 267--273, 1982.

\bibitem{Saegusa2009}
R.~Saegusa, G.~Metta, G.~Sandini, and S.~Sakka, ``Active motor babbling for
  sensorimotor learning,'' in \emph{{Int. Conf. Robotics and Biomimetics}}, Feb
  2009, pp. 794--799.

\bibitem{kohonen2013essentials}
T.~Kohonen, ``Essentials of the self-organizing map,'' \emph{Neural networks},
  vol.~37, pp. 52--65, 2013.

\bibitem{collins2010neuron}
C.~E. Collins, D.~C. Airey, N.~A. Young, D.~B. Leitch, and J.~H. Kaas, ``Neuron
  densities vary across and within cortical areas in primates,''
  \emph{Proceedings of the National Academy of Sciences}, vol. 107, no.~36, pp.
  15\,927--15\,932, 2010.

\bibitem{young2013cell}
N.~A. Young, C.~E. Collins, and J.~H. Kaas, ``Cell and neuron densities in the
  primary motor cortex of primates,'' \emph{Frontiers in neural circuits},
  vol.~7, p.~30, 2013.

\bibitem{fritzke1995growing}
B.~Fritzke, ``A growing neural gas network learns topologies,'' in
  \emph{Advances in neural information processing systems}, 1995, pp. 625--632.

\bibitem{Journals:Kohler1962}
I.~Kohler, ``Experiments with goggles,'' \emph{Scientific American}, vol. 206,
  no.~5, pp. 62--73, 1962.

\bibitem{tensorflow2015-whitepaper}
\BIBentryALTinterwordspacing
M.~Abadi \emph{et~al.}, ``{TensorFlow}: Large-scale machine learning on
  heterogeneous systems,'' 2015, software available from tensorflow.org.
  [Online]. Available: \url{http://tensorflow.org/}
\BIBentrySTDinterwordspacing

\end{thebibliography}
\bibliographystyle{IEEEtran}

\end{document}